\documentclass[sigconf]{acmart}

\usepackage{algorithm}
\usepackage{algorithmic}
\usepackage{multirow}

\AtBeginDocument{%
  }

\copyrightyear{2025}
\acmYear{2025}
\setcopyright{acmlicensed}\acmConference[MM '25]{Proceedings of the 33rd ACM International Conference on Multimedia}{October 27--31, 2025}{Dublin, Ireland}
\acmBooktitle{Proceedings of the 33rd ACM International Conference on Multimedia (MM '25), October 27--31, 2025, Dublin, Ireland}
\acmDOI{10.1145/3746027.3755546}
\acmISBN{979-8-4007-2035-2/2025/10}

\begin{document}

\title{TopoImages: Incorporating Local Topology Encoding into \\Deep Learning Models for Medical Image Classification}

\author{Pengfei Gu}
\affiliation{%
  \institution{University of Texas Rio Grande Valley}
  \city{Edinburg}
  \state{Texas}
  \country{USA}
}
\email{pengfei.gu01@utrgv.edu}

\author{Hongxiao Wang}
\affiliation{%
  \institution{Capital Normal University}
  \city{Beijing}
  \state{}
  \country{China}
}
\email{hxwang1992@gmail.com}

\author{Yejia Zhang}
\affiliation{%
  \institution{University of Notre Dame}
  \city{Notre Dame}
  \state{Indiana}
  \country{USA}
}
\email{chazhang0310@gmail.com}

\author{Huimin Li}
\affiliation{%
  \institution{The University of Texas at Dallas}
  \city{Richardson}
  \state{Texas}
  \country{USA}
}
\email{huimin.li@utdallas.edu}

\author{Chaoli Wang}
\affiliation{%
  \institution{University of Notre Dame}
  \city{Notre Dame}
  \state{Indiana}
  \country{USA}
}
\email{chaoli.wang@nd.edu}

\author{Danny Chen}
\affiliation{%
  \institution{University of Notre Dame}
  \city{Notre Dame}
  \state{Indiana}
  \country{USA}
}
\email{dchen@nd.edu}

\renewcommand{\shortauthors}{Gu et al.}

\begin{abstract}
Topological structures in image data, such as connected components and loops, play a crucial role in understanding image content (e.g., biomedical objects).
Despite remarkable successes of numerous image processing methods that rely on appearance information, these methods often lack sensitivity to topological structures when used in general deep learning (DL) frameworks.
In this paper, we introduce a new general approach, called TopoImages (for Topology Images), which computes a new representation of input images by encoding local topology of patches. 
In TopoImages, we leverage persistent homology (PH) to encode geometric and topological features inherent in image patches. 
Our main objective is to capture topological information in local patches of an input image into a vectorized form.
Specifically, we first compute persistence diagrams (PDs) of the patches,
and then vectorize and arrange these PDs into long vectors for pixels of the patches.
The resulting multi-channel image-form representation is called a TopoImage.
TopoImages offers a new perspective for data analysis. 
To garner diverse and significant topological features in image data and ensure a more comprehensive and enriched representation, we further generate multiple TopoImages of the input image using various filtration functions, which we call multi-view TopoImages.
The multi-view TopoImages are fused with the input image for DL-based classification, with considerable improvement.
Our TopoImages approach is highly versatile and can be seamlessly integrated into common DL frameworks. Experiments on three public medical image classification datasets demonstrate noticeably improved accuracy over state-of-the-art methods.
\end{abstract}






\maketitle

\section{Introduction}
\label{sec:intro}
\begin{figure}[t!]
    \centering
    \includegraphics[width=0.88\columnwidth]{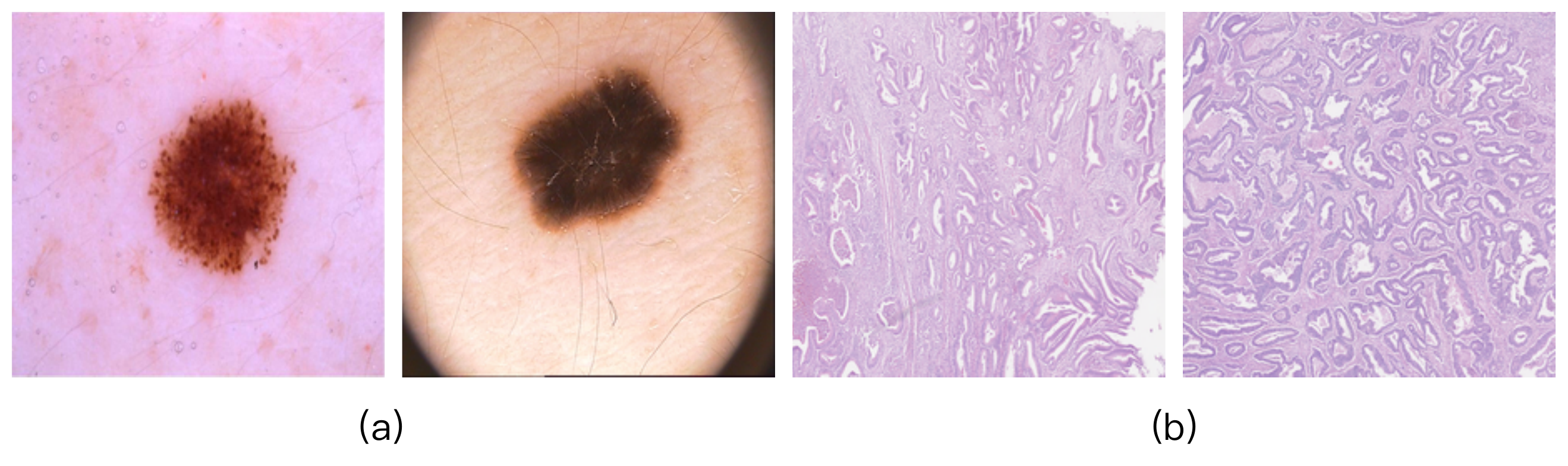}
    \caption{Image examples of the ISIC 2017 dataset~\cite{codella2018skin} (a) and the ExtCRC dataset~\cite{shaban2020context} (b), showing geometric information (e.g., boundary points/curves of a skin lesion or a gland) of the images.} 
    \label{fig1}
\end{figure}

Images contain rich information on object appearance and geometry.
Object geometry information often includes object shape, texture, topology, and instance distribution. For example, images of the ISIC 2017 skin lesion classification dataset show the numbers and shapes of melanomas~\cite{codella2018skin}. The extended colorectal cancer grading dataset provides images encompassing the shapes of glands~\cite{shaban2020context} (e.g., see Fig.~\ref{fig1}).
While analyzing images based on object appearance information (e.g., pixel intensities) is ``natural'' and fruitful, it is also highly beneficial to explicitly exploit object geometry information (e.g., object distribution patterns), which will enable a much richer extraction of image content. Medical image datasets often contain especially distinct and useful geometric/topological patterns due to the relatively high stability, regularity, or commonality in their imaging modalities, imaging settings, target objects, etc. (e.g., MR images of bones and cartilages in joint structures \cite{Peng-KCB-2022}, micro-CT images of cartilages 
in embryonic chondrocranium \cite{Perrine-eLife-2022}, fluorescence microscopy images of cells, tissues, and tumors in the brain \cite{Guldner-Cell-2022}). For example, in the knee joint anatomy \cite{Peng-KCB-2022}, 
target bones and cartilages are normally arranged in relatively stable geometric positions with some possible variations of shapes, and such ``stable'' topological patterns of the knee structure in images can be beneficially exploited for disease assessment and analysis.

Recently, Topological Data Analysis (TDA) has emerged as a powerful methodology for image analysis \cite{edelsbrunner2002topological}, offering effective methods such as persistence landscape~\cite{bubenik2015statistical}, persistence scale-space kernel~\cite{reininghaus2015stable}, and persistence images (PIs)~\cite{adams2017persistence}.
Persistent homology (PH), a common tool of TDA, tracks how topological patterns appear and disappear as a nested sequence in a topological space. 
A prominent representation of PH is persistence diagram (PD), which encodes topological information of input data (e.g., an image, a point set, or a graph) as a set of points. These points can then be vectorized for feature representation.
PH-based methods offer new insights into data analysis due to their important properties: (1) They are invariant to continuous object deformation and scaling, providing robustness against perturbation and noise, and (2) they are versatile and adaptable, making them applicable to a wide variety of data types and objects.
Deep learning (DL) has become a prevalent technique for image data analysis, 
capturing appearance-based information on fixed pixel grids.
For instance, convolutional neural networks (CNNs) \cite{ronneberger2015u,gu2021k} utilize learned convolution kernels, while Transformers \cite{dosovitskiy2020image,gu2023convformer} rely on self-attention mechanisms between regional patches in images.
Although geometric information in images can be jointly learned, such appearance-based methods have several key drawbacks: (1) The fixed learned kernels in DL models lack flexibility to accommodate elastic deformation of objects, and (2) DL methods are not adequately sensitive to topological structures of objects, such as connected components and loops.

We hypothesize that an effective representation of topological information, especially local topology information, can complement common appearance-based representations for medical image analysis.
Hence, we propose a new general approach, called TopoImages (for Topology Images), which computes a new representation of input images by encoding the local topology of patches into a vectorized form.
%
%
Specifically, we first compute and vectorize the PDs of the local patches in an input image $I$, applying the method of PIs~\cite{adams2017persistence}.
The vectorized PDs are then reshaped into long vectors for each pixel of the patches, and the long vectors for all the pixels of $I$ are put together into a multi-channel image-form representation (TopoImage) of the input image.
This TopoImage enables us to represent the input image with specific local topological information, offering a general topological representation that is very different from appearance-based representations. 
Our TopoImages approach is highly versatile and can be seamlessly integrated into common DL frameworks for applications.

To capture diverse and significant topological features in image data and ensure a more comprehensive and enriched representation, we further propose to generate multiple TopoImages of an input image using various filtration functions, which are called multi-view TopoImages.
We develop a new fusion module to complement the input image with the multi-view TopoImages and integrate them into existing DL frameworks.
Specifically, we first apply convolution operations to compress the multi-view TopoImages to the same number of channels as the input image.
Next, we fuse the compressed multi-view TopoImages with the input image via an addition operation. 
Finally, the composite (fused) image is 
fed to a DL model for an application task.
The performance of multi-view TopoImages is evaluated using common DL networks for medical image classification tasks
%
on three public datasets. The results demonstrate the effectiveness of our new approach, yielding considerable improvements 
over state-of-the-art (SOTA) methods.

Our main contributions are summarized as follows. (1) We propose a general new local topology-based representation for input images, which encodes specific local topological information into an image form, called TopoImages. (2) We further generate multi-view TopoImages of input images with different filtration functions, and introduce a new fusion module to complement the input images with multi-view TopoImages. (3) We demonstrate the effectiveness of our TopoImages approach in conjunction with common DL networks on medical image classification tasks.

\section{Related Work}\label{sec:related-work}

\subsection{TDA for Image Analysis}
\label{sec:rw:1}
Compared to common CNN methods, TDA offers several distinguished advantages in processing images.
First, TDA exhibits greater robustness to noise caused by imaging conditions and artifacts, enabling the identification of prominent topological structures.
%
For instance, loop topology can be used to enhance the completeness of cell membrane structures in noisy image background~\cite{hu2019topology, hu2020topology, he2023toposeg}. With 
Morse cancellation, road networks can be reconstructed from noisy GPS tracking records~\cite{dey2017improved}. The complex fiber topology of neurons can help their reconstruction in microscopy images~\cite{banerjee2020semantic}. 
Second, TDA is quite sensitive to topological structures that are correlated with medical data. 
For example, the loop topology of glands is closely associated with the Gleason score in prostate cancer diagnosis~\cite{lawson2019persistent}. The changes in topology on skin lesion textures and in the distribution of cells in breast pathology images can be used to enhance disease classification accuracy~\cite{peng2024phg}. Compared to CNNs, these topology-based methods are more acceptable to physicians as they provide more interpretable and comprehensible evidence.
However, the known TDA-based methods mostly pay attention only to \emph{global} topological features and generate just {\it one} persistence diagram for a whole image. In contrast, we focus on \emph{local} topology of patches around each pixel and thus consider many PDs for an image, as these PDs provide finer-grained region-level local topology to complement pixel-level appearance information in the whole image.

\subsection{PH Representations and Vectorization}
\label{sec:rw:2}
Finding an effective PH representation is fundamental for downstream topological analysis tasks.
Non-vectorization topological analysis methods are more direct.
For instance, PD and persistent barcode plot pairs (birth time, death time) of topological structures in a coordinate system~\cite{edelsbrunner2002topological}. One can measure the topological distance between two objects by computing the 1-Wasserstein distance between their PDs~\cite{aukerman2022persistent}.
The merge tree uses a hierarchical tree structure to represent the evolution of topological structures as the filter function value changes~\cite{curry2022decorated}.
However, these methods pose challenges to machine learning methods, including DL algorithms, which require vector representations. 
To address this problem, one technique, persistence landscape~\cite{bubenik2015statistical}, represents PH as functions that can be studied within a vector space. 
Another interesting method is PIs~\cite{adams2017persistence}, which maps a PD to a finite-dimensional vector representation space that has been proven to be stable against small input perturbations. 
In this work, instead of proposing a new PH vectorization method for DL frameworks, we use PIs~\cite{adams2017persistence} to enrich input images with local topology-encoded vector representations. In this way, both appearance information and topological information in medical images can be directly combined and processed by DL frameworks.

\subsection{Integrating Topological Information into DL Frameworks}
\label{sec:rw:3}
Known data analysis methods for incorporating topological information into DL frameworks can be mainly categorized as follows. (1) Regulation: These methods focus on training a common DL framework by designing a topology-based loss function in order to encourage geometric correctness. 
For instance, the Betti number was used to regulate segmentation~\cite{clough2020topological,gu2024self,adame2025topo}; the number of branches was exploited to enhance continuity in retinal vessel segmentation~\cite{shit2021cldice}. 
(2) Guided information aggregation: In~\cite{wang2023ccf}, a hierarchical merge tree structure was used to guide DL information aggregation from cell level to cell cluster (or cell community) level.
(3) Data representation: These methods use topological representations, such as PIs, as secondary input to enhance DL performance. 
For instance, multi-modal datasets were mapped as multi-channel PIs, creating a unified feature space for downstream analysis~\cite{myers2023topfusion}.
In~\cite{du2022distilling}, PDs were directly employed to distill knowledge for DL model output.
Our TopoImages fall in the data representation category. Compared to methods in the other categories, data representation methods are more generalizable and can be easily integrated into existing DL frameworks. Notably, our TopoImages can be directly input into existing DL frameworks without requiring any modifications to their architectures.

\section{Preliminaries}\label{subsec:Preliminaries} 


\subsection{Persistence Diagrams}
\begin{figure}[t!]
    \centering
    \includegraphics[width=0.85\columnwidth]{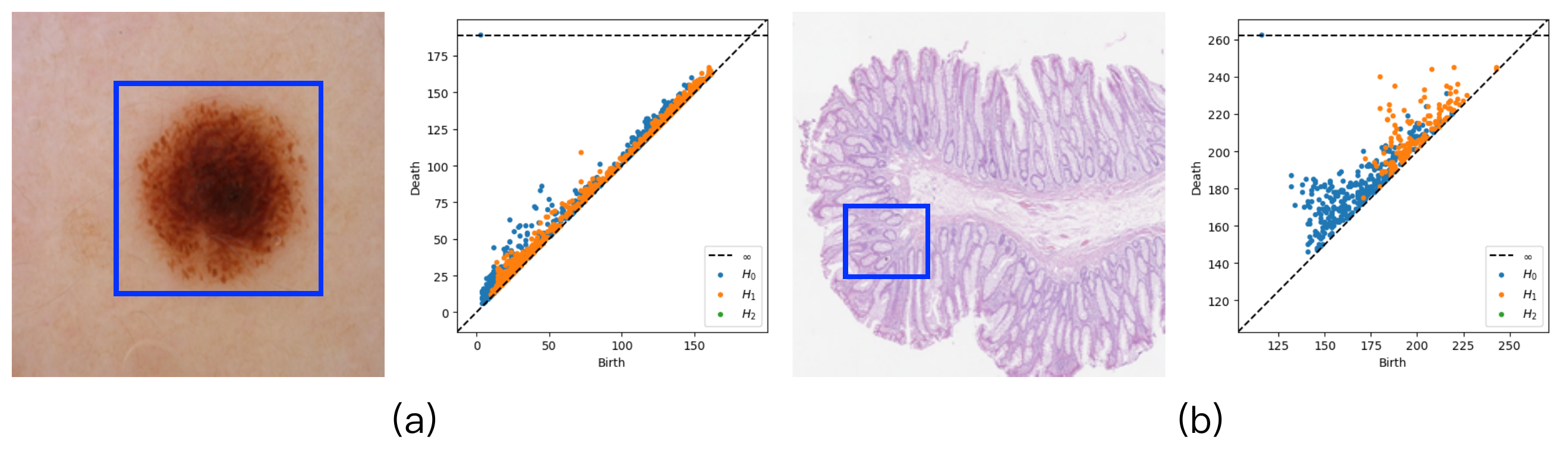}
    \caption{Examples of persistence diagrams of patches (indicated by blue boxes) in images of the ISIC 2017 dataset (the right image in (a)) and ExtCRC dataset (the right image in (b)). In these PDs, the blue points represent the 0-D persistent homology (H0), and the orange points represent the 1-D persistent homology (H1).} 
    \label{fig:pds}
\end{figure}

A common topological descriptor derived from PH is PD. 
This descriptor provides a (visual) summary of births and deaths of topological events. 
For example, it tracks when a loop appears and
disappears as the complexity of data grows (e.g., while the filtration parameter value increases). 
PD is a multi-set of points typically plotted in the plane, where each point represents a topological feature (e.g., a connected component or a hole) of a dataset. 
More specifically, a PD is a set of points in $R^2$; each topological feature has a birth time $b$ and a death time $d$, and is represented in a diagram as a point 
$(b,d)$.
Fig.~\ref{fig:pds} shows some examples of 
PDs of patches in images from the ISIC 2017 skin lesion classification dataset~\cite{codella2018skin} and the extended colorectal cancer grading dataset~\cite{shaban2020context}. 
However, PDs can exhibit a complicated structure, and are difficult to integrate directly into machine learning workflows.

\begin{figure*}[t!]
    \centering
    \includegraphics[width=0.85\textwidth]{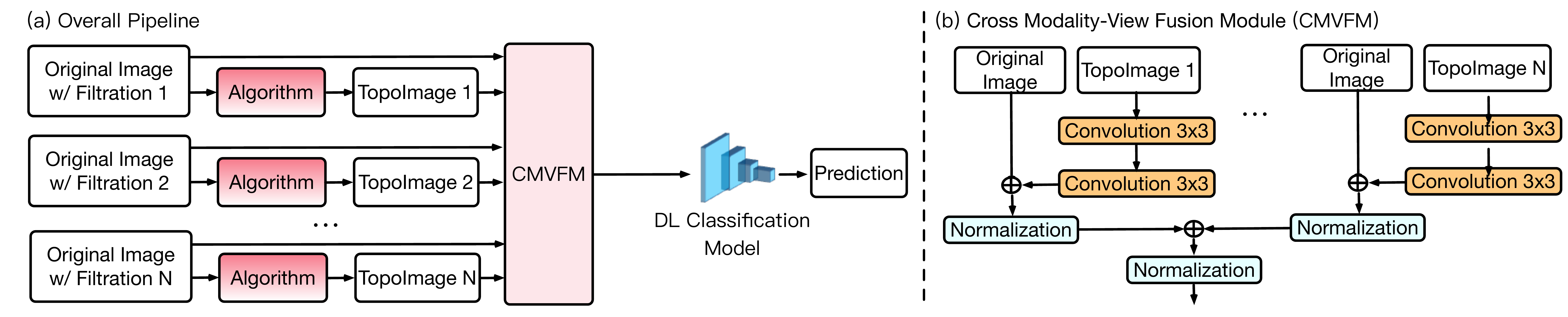}
    \caption{(a) Illustrating the overall workflow of our approach. Given an input image with different filtration functions, Algorithm~\ref{algorithm1} is first used to compute the corresponding multi-view TopoImages. A new cross-modality-view fusion module (CMVFM) is proposed to complement the input image with the multi-view TopoImages. The fused image is then fed to an existing DL model (e.g., ConvNeXt~\cite{liu2022convnet}) to generate the final prediction.
    (b) The structure of our proposed CMVFM. For simplicity, the batch normalization and rectified linear unit that follow each convolution layer are omitted.
    } 
    \label{fig:pipeline}
\end{figure*}

\subsection{Persistence Images}
\label{subsec-PIs}
PIs~\cite{adams2017persistence} are a stable method for vectorizing topological features in a summarized PD for DL-based applications. 
%
PIs are formed by overlaying distributions centered on each point in a PD, summing these distributions to form a surface, and then discretizing the surface to build an image.

In~\cite{adams2017persistence}, there are three key parameter choices for generating a PI: the resolution, distribution, and weighting function.
The {\it resolution} of the PI corresponds to a size $H'\times W'$ grid overlay on the PD, which represents the number of pixels along each dimension of the PI.
The {\it distribution} is often a Gaussian function.
The {\it weighting function}, which maps persistence pairs to weights, is usually a simple linear function. We will follow these definitions when describing our proposed method in Section~\ref{sec:Methods}.

PIs preserve essential topological information and provide a robust and stable representation against perturbations in the data. 
This property is particularly valuable 
when data are subject to noise and variability (e.g., in image analysis and signal processing).
The vectorized nature of PIs enables their integration into machine learning methods.
However, due to their relatively low resolution (e.g., 7$\times$7), it is difficult to integrate them effectively into existing DL networks.

\section{Our Method}
\label{sec:Methods}
Fig.~\ref{fig:pipeline}(a) shows the overall workflow of our 
approach.
Generally, given an input image $I$ with different filtration functions, the corresponding multi-view TopoImages of $I$ are first computed using Algorithm~\ref{algorithm1}. A cross-modality-view fusion module (CMVFM) is then applied to complement the input image with these multi-view TopoImages, which are subsequently fed to existing DL models for specific tasks.
%

%

\subsection{PH of Images}\label{subsec:topo}
Given an input image $I$, we represent it with a \textit{cubical complex} $C$, which
typically takes each pixel of $I$ as a vertex and contains connectivity information on vertex neighborhoods via edges, squares, and their higher-dimensional counterparts~\cite{wagner2011efficient,kaji2020cubical}.
PH~\cite{edelsbrunner2002topological} is a powerful mathematical tool to extract topological features of different dimensions from a cubical complex (e.g., connected components ($0$-D) and cycles/loops ($1$-D)). 
It combines the homology of super-level sets by sweeping a threshold through the entire set of real numbers.
Specifically, for a threshold value $\tau \in \mathbb{R}$, a cubical complex is defined as: $C^{(\tau)} := \{x\in I \ | \ f(x) \geq \tau\},$
where a filtration function $f(x)$ maps a pixel $x$ to its value (e.g., its intensity value). 
When sweeping the threshold, the topology changes only at a finite number of values, $\tau_1\geq \tau_2 \geq \cdots \geq \tau_{m-1} \geq \tau_m$, and we obtain a sequence of nested cubical complexes: $\emptyset  \subseteq C^{(\tau_1)} \subseteq C^{(\tau_2)} \subseteq \cdots \subseteq C^{(\tau_{m-1})}\subseteq C^{(\tau_m)} = I,$
which is the \textit{super-level set filtration}.
PH tracks topological features across all the cubical complexes in this filtration, representing each topological feature as a tuple $(\tau_i, \tau_j)$, with $\tau_i \geq \tau_j,$ indicating the cubical complexes in which the feature appears and disappears. 
For example, a $0$-D tuple $(\tau_i, \tau_j)$ represents a connected component that appears at threshold value $\tau_i$ and disappears at threshold value $\tau_j$;
$\tau_j -\tau_i$ represents the persistence (lifespan) of that topological feature. 
The tuples of all the $k$-D ($k\in\{0,1\}$) topological features are kept in the \textit{$k$-th} PD, $D_I^{k}$, which is a multi-scale shape descriptor of all the topological features of image $I$.

We employ the Cubical Ripser software~\cite{kaji2020cubical} to compute the PDs of input images in our experiments. We choose this software since it is fast and memory-efficient in computing PDs of weighted cubical complexes.

\begin{figure*}[t!]
    \centering
    \includegraphics[width=0.85\textwidth]{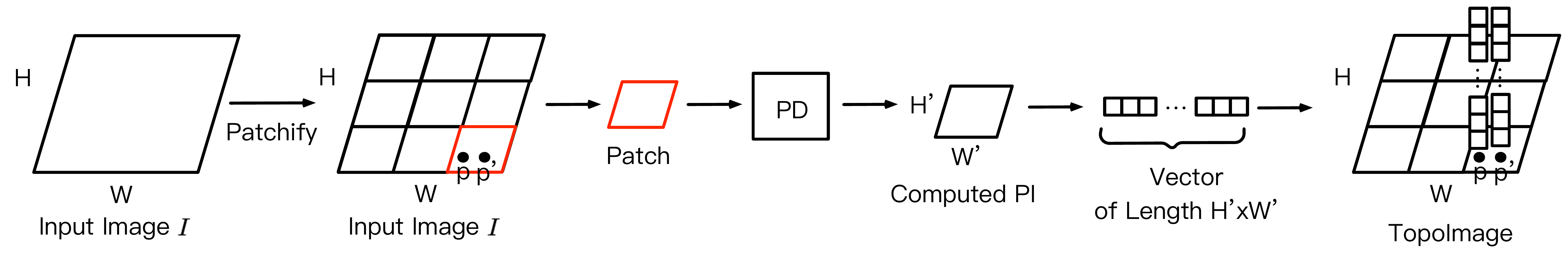}
    \caption{Illustrating the pipeline of our Algorithm~\ref{algorithm1} for computing a TopoImage. Given a size $H \times W$ (e.g., $224 \times 224$) input image $I$, we first divide $I$ into uniform patches (e.g., size $28 \times 28$ each). Then, the PD of each patch of $I$ is computed, and a size (or resolution) $H' \times W'$ (e.g., $7 \times 7$) persistence image (PI) is computed for the PD of the patch. Next, the PI is transformed into a long vector of length $H'\times W'$, which is then assigned to each of the pixel points of the corresponding patch in the TopoImage being constructed. For simplicity, the computation is shown for only two pixels $p$ and $p'$ in one patch of $I$.} 
    \label{fig:algo}
\end{figure*}

\subsection{TopoImages: Representing Images with Local Topology Encoded Vector Representations}\label{subsec:TopoImages}
Since PD is a commonly-used representation of PH, many methods exist for mapping PDs into vector representations that can be used by machine learning frameworks.
For example, Cubical Ripser~\cite{kaji2020cubical} introduced lifetime-enhanced images that incorporate PH information into input images, which are then fed to DL networks for classification (e.g., on the MNIST dataset \cite{MNIST-LeCun}).
%
But, the lifetime-enhanced images are often very sparse, containing many zeros, and may lose certain key information on the PDs. 
PIs~\cite{adams2017persistence} are a stable method for vectorizing topological features summarized in PDs for DL model usage.
This method can be naturally integrated into DL networks but still has some drawbacks:
(1) Modifications or rebuilding of DL architectures are required for direct use of PIs due to their lower resolution  (e.g., 7$\times$7) compared to that of the input images;
(2) local topology information is not well explored by PIs.

To address these issues, we develop the TopoImages approach, which computes a new representation of an input image by encoding local topological features of its patches.
Our main idea is to map local topological information of a patch into a long vector (e.g., of a length $H' \times W'$) for representing the PD of the pixels in the patch.

Fig.~\ref{fig:algo} shows our pipeline for computing one patch of a TopoImage.
Specifically, for a size $H \times W$ (e.g., $224 \times 224$) input image $I$, we divide $I$ into uniform patches (e.g.,  size $28 \times 28$ each).
Then, the PD of each patch $P$, $P\!D(P)$, is computed, and a size $H' \times W'$ (e.g., $7 \times 7$) PI, $P\!I(P)$, is computed for $P\!D(P)$. 
$P\!I(P)$ is then transformed into a long vector of length $H'\times W'$ (e.g., $7\times 7$ = 49), which is assigned to each of the pixel points of patch $P$ in the TopoImage that we are building.
After processing all the patches of $I$ in this way, a multi-channel ``image'' $I'$ is built (by Algorithm~\ref{algorithm1}, not by an imaging modality).
Each ``pixel'' in patch $P$ of $I'$ is associated with the same long vector computed from $P\!I(P)$ in $I$, and the ``pixels'' in different patches of $I'$ receive different long vectors. 
The number of channels of $I'$ is equal to the length of such a long vector, which is the height $\times$ width product of $P\!I(P)$ ($H'\times W'$).
The multi-channel ``image'' $I'$ thus constructed is a TopoImage of $I$, whose height and width are the same as those of $I$ ($H \times W$).

Algorithm~\ref{algorithm1} gives the procedure for creating a TopoImage of an input image using a given filtration function.
%
%
%
%

%
We should note that selecting a suitable patch size is critical for effectively capturing the local topology. To achieve this, we use SAM~\cite{kirillov2023segment} to produce segmentation masks in the training images of the dataset, and approximate the average object size by averaging the sizes of SAM segmentation masks.
Among the patch sizes \{7×7, 14×14, 28×28, 56×56, 112×112\}, we select the one that closely aligns with the approximate average object size as the patch size for dividing the input images.
We investigate the effect of patch size choices experimentally in Section~\ref{sub:patch-size}.

%

\begin{algorithm}[t]
\caption{Computing a TopoImage}\label{algorithm1}
\begin{algorithmic}[1]
    \REQUIRE An input image $I$ of size $H \times W$
    \ENSURE A TopoImage of $I$
    \STATE Load the input image $I$ in floating-point representation
    \STATE Let $I'$ be an empty multi-channel ``image'' of size $H \times W$, with its number of channels equal to the height $\times$ width product ($H' \times W'$) of the persistence image (PI)
    \STATE Divide the input image $I$ into uniform patches
    \FOR{each patch $P$ of image $I$}
        \STATE Normalize all the pixel values (e.g., intensities) of $P$ to floating-point numbers in the range [0, 1]
        \STATE Compute the PD of the normalized patch
        \STATE Vectorize the PD using the PIs method
        \STATE Arrange the vectorized PD as a long vector, and assign the vector to all the pixel points of the corresponding patch $P$ in the multi-channel image $I'$
    \ENDFOR
    \RETURN the multi-channel image $I'$
\end{algorithmic}
\end{algorithm}


\subsection{Complementing Input Images with Multi-View TopoImages for DL Models}\label{subsec:combine}
Medical images often exhibit variability due to differences in imaging modalities, patient variations, and pathological conditions. Different filtration functions can provide diverse perspectives on the topological properties of the image data. For example, intensity filtration may differentiate areas of varying brightness using intensity values, while gradient filtration may highlight the edges of objects, making it easier to delineate organs such as livers and kidneys.

To capture diverse and significant topological features in images and ensure a more comprehensive and enriched representation, we further use Algorithm~\ref{algorithm1} to compute multiple TopoImages of an input image with different filtration functions, called multi-view TopoImages.
Multi-view TopoImages offers a few advantages. (1) They preserve essential diverse local topological information while providing a robust and stable vector representation, making them invariant to deformation and scaling. This 
makes multi-view TopoImages applicable to a wide variety of data types and objects. 
(2) Their vectorized form enables them to be seamlessly integrated into common DL frameworks. More importantly, they have the same formation (i.e., height and width) as the input image, allowing them to be directly input into existing DL models without modifying their architectures.
Due to these advantages, our multi-view TopoImages can be used to effectively complement traditional appearance-based image representations, enhancing the performance of known DL models for their 
tasks.

To complement an input image $I$ with multi-view TopoImages for existing DL networks, a straightforward way is to concatenate the multi-view TopoImages with $I$.
But, this may cause some issues. (1) TopoImages, having a much larger number of channels than that of $I$, can dominate the input image in DL model learning. 
(2) The features captured by different filtration functions may not be well aligned or normalized, giving discrepancies in the feature scales.
To address these issues, we devise a new cross-modality-view fusion module (CMVFM). As shown in Fig.~\ref{fig:pipeline}(b), CMVFM first compresses each multi-view TopoImage to the same number of channels as $I$.
This compression is attained with two convolutional blocks, each of which consists of a convolution layer with a 3$\times$3 kernel size, a batch normalization, and a rectified linear unit layer.
We then fuse the compressed TopoImage with $I$ via an addition operation. 
Next, we normalize the fused images to the range [0,1].
Finally, we add the normalized images together and normalize the resulting image to the range [0,1].
This approach allows that the resulting image can be directly input into existing (pre-trained) DL networks without any architecture modifications.
Fig.~\ref{fig:pipeline}(a) shows the main steps of our overall method. 
%
Clearly, this method is also applicable to an input image with a single filtration function, by simply complementing the image with the corresponding TopoImage, without the need to add multiple fused images and normalize the sum image to the range [0,1].

\begin{figure}[t!]
    \centering
    \includegraphics[width=0.88\columnwidth]{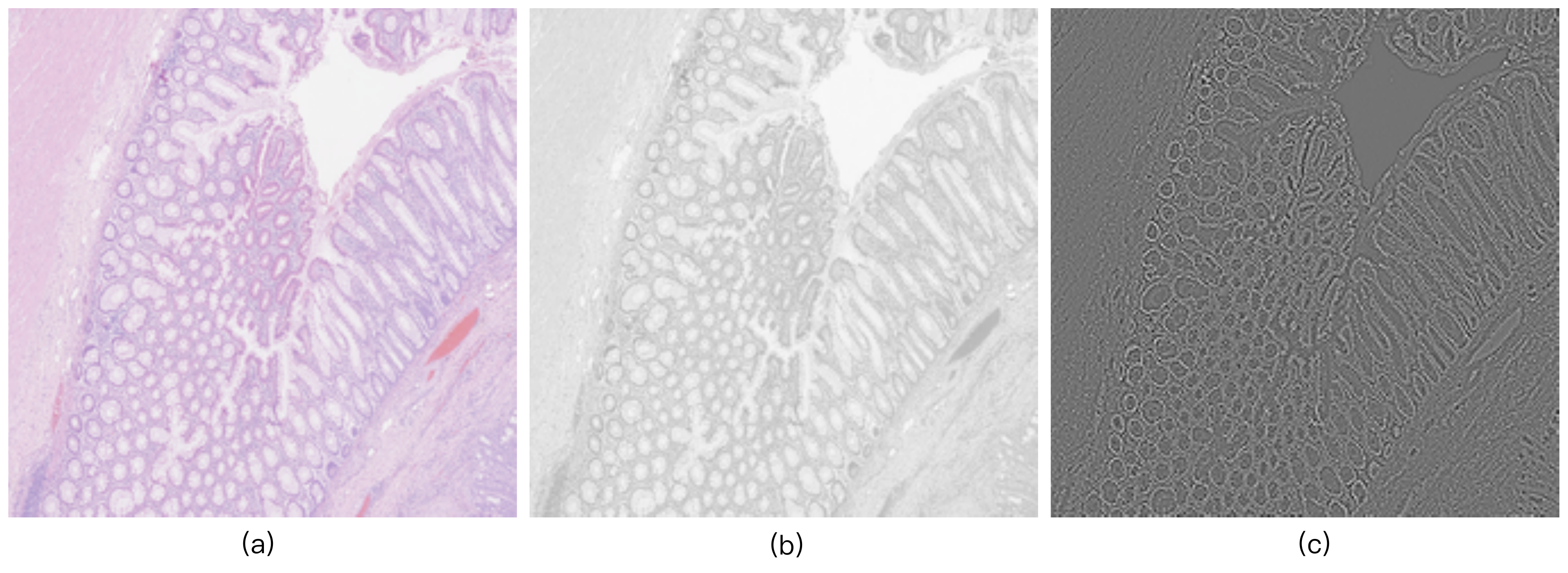}
    \caption{Image examples of the ExtCRC dataset~\cite{shaban2020context} used to create sublevel filtrations. (a) An original image; (b) the image with magnitude values; (c) the image with gradient magnitude values.
    } 
    \label{fig-filtrations}
\end{figure}

In this work, when computing PDs, we use the intensity and gradient filtration functions. Specifically, for grayscale images, we utilize pixel intensity values for the intensity filtration function, and gradient magnitude values computed by the Laplacian operator for the gradient filtration function.
For color images, we use the magnitude values of the three RGB channels of the pixels for the intensity filtration function, and the average gradient magnitude values computed by the Laplacian operator across the three RGB channels for the gradient filtration function (see examples of Fig.~\ref{fig-filtrations}).

We chose these two filtration functions because: (i) Intensity filtrations capture pixel-value-based topological structures (like connected bright/dark regions) by thresholding images at different intensity levels. This approach reveals features corresponding to coherent intensity regions and is particularly effective for images with clear contrast. Intensity filtrations preserve the natural hierarchy of image features based on their brightness values.
(ii) Gradient filtrations capture topological features around edges or boundaries by using the magnitude of intensity changes. This method emphasizes transitions between regions, making it less sensitive to absolute intensity values and better at capturing shape information independent of illumination variations. These features are often meaningful in medical images (e.g., boundary loops around lesions or glands).
Together, these complementary filtrations capture different aspects of image structure.

%

\section{Experiments and Results}
\label{sec:Experiments and Results}

\subsection{Datasets and Experimental Setup}\label{datasets}
We evaluate TopoImages on three public medical image classification datasets: 
(1) The ISIC 2017 skin lesion classification dataset~\cite{codella2018skin} (ISIC 2017): 
It contains 2000 training, 150 validation, and 600 test images. Our experiments focus on task-3A: melanoma detection.
(2) The curated breast imaging subset of a digital database for screening mammography dataset (CBIS-DDSM):
This dataset~\cite{lee2017curated} contains 1566 participants and 6775 studies, which are categorized as benign or malignant. We use the data split given by the dataset provider: 20\% 
for testing and 80\% for training.
(3) The extended colorectal cancer grading dataset (ExtCRC):
This dataset~\cite{shaban2020context} contains 300 H\&E-stained colorectal cancer subtyping pathology images, which are divided into three categories (Grades 1, 2, and 3).
We randomly split 80\% of the data for training and 20\% for testing.
We resize all images of each dataset to 224$\times$224. For the ISIC 2017 and CBIS-DDSM datasets, our experiments conduct three runs using different seeds. For the ExtCRC dataset, we perform random data splitting for three times, reporting the average results of the experiments.
Two common evaluation metrics are used: accuracy (Acc) and area under the receiver operating characteristic
 curve (AUC).

%

\begin{table}[t]
    \centering
    \caption{Results of various methods on the ISIC 2017 dataset. The best results are marked in {\bf bold}, and the second-best results are \underline{underlined}. The same goes for the other tables.
    }
    \label{tab:isic17}
    \scalebox{0.8}{%
    \begin{tabular}{l|l|l}
        \hline
        Method &AUC  ($\uparrow$)  &Acc ($\uparrow$) \\\hline\hline
        Galdran et al.~\cite{galdran2017data} &76.50	&	 48.00\\\hline
        Vasconcelos et al.~\cite{vasconcelos2017increasing} &79.10 	&83.00	\\\hline
        D{\'\i}az~\cite{diaz2017incorporating} &{85.60} 	&	82.30\\\hline
        Zhang et al.~\cite{zhang2019medical}  &83.00	&83.00	\\\hline
         ARDT-DenseNet~\cite{wu2020skin} &\textbf{87.90} 	&\underline{86.80}	\\\hline
        Suraj et al.~\cite{mishra2022data} &83.10 	&{86.00}	\\\hline
         Gu et al.~\cite{gu2024boosting} &83.27 	&{86.67}	\\
       \hline
        \hline
        ResNet50~\cite{he2016deep}&	79.01  &  82.67	  \\
       \hline
       \begin{tabular}[c]{@{}l@{}} ResNet50 + PIs~\cite{adams2017persistence}\end{tabular}
      &80.08	  &  82.33	  \\
       \hline
       \begin{tabular}[c]{@{}l@{}} ResNet50 + LTEIs~\cite{kaji2020cubical}\end{tabular}
        &	76.21  & 78.83 	  \\\hline
        \begin{tabular}[c]{@{}l@{}} ResNet50 + TopoImage w/ intensity filtration\end{tabular}
        &80.11	  &82.00  	  \\\hline
         \begin{tabular}[c]{@{}l@{}} ResNet50 + TopoImage w/ gradient filtration\end{tabular}
        &80.43	  & 83.00 	  \\\hline
        \begin{tabular}[c]{@{}l@{}} ResNet50 + multi-view TopoImags \\ w/ intensity \& gradient filtrations (ours)\end{tabular}
        &81.47	  &  83.33	  \\\hline\hline
       ConvNeXt~\cite{liu2022convnet}&	83.16  &  85.00	  \\
       \hline
       \begin{tabular}[c]{@{}l@{}} ConvNeXt + PIs\end{tabular}
      &	81.15  &82.83\\
       \hline
       \begin{tabular}[c]{@{}l@{}} ConvNeXt + LTEIs\end{tabular}
        &	78.83  & 79.50	  \\\hline
        \begin{tabular}[c]{@{}l@{}} ConvNeXt + TopoImage w/ intensity filtration\end{tabular}
        &83.74	  &  84.00	  \\\hline
         \begin{tabular}[c]{@{}l@{}} ConvNeXt + TopoImage w/ gradient filtration\end{tabular}
        &84.09	  &  85.50	  \\\hline
        \begin{tabular}[c]{@{}l@{}} ConvNeXt + multi-view TopoImags \\ w/ intensity \& gradient filtrations (ours)\end{tabular}
        &85.59	  &  86.00	  \\\hline\hline
         Swin-Transformer v2~\cite{liu2022swin}&80.78	  & 83.50 	 \\
       \hline
       \begin{tabular}[c]{@{}l@{}} Swin-Transformer v2 + PIs\end{tabular}
      &79.90	  &  80.67	  \\
       \hline
       \begin{tabular}[c]{@{}l@{}} Swin-Transformer v2 + LTEIs\end{tabular}
        &	74.89  &  79.83	  \\\hline
        \begin{tabular}[c]{@{}l@{}} Swin-Transformer v2 + TopoImage w/ intensity filtration\end{tabular}
        &	81.38  & 83.67 	  \\\hline
         \begin{tabular}[c]{@{}l@{}} Swin-Transformer v2 + TopoImage w/ gradient filtration\end{tabular}
        &	81.93  & 84.00 	  \\\hline
        \begin{tabular}[c]{@{}l@{}} Swin-Transformer v2 + multi-view TopoImags \\ w/ intensity \& gradient filtrations (ours)\end{tabular}
        &	82.71  &  84.67	  \\\hline\hline
         PHG-Net~\cite{peng2024phg}&84.14	  &  85.17	  \\
       \hline
       \begin{tabular}[c]{@{}l@{}} PHG-Net + PIs\end{tabular}
      &83.60	  &  84.80	  \\
       \hline
       \begin{tabular}[c]{@{}l@{}} PHG-Net + LTEIs\end{tabular}
        &79.47	  &  80.83	  \\\hline
        \begin{tabular}[c]{@{}l@{}} PHG-Net + TopoImage w/ intensity filtration\end{tabular}
        &85.02	  & 84.50 	  \\\hline
         \begin{tabular}[c]{@{}l@{}} PHG-Net + TopoImage w/ gradient filtration\end{tabular}
        &85.58	  & 85.83 	  \\\hline
        \begin{tabular}[c]{@{}l@{}} PHG-Net + multi-view TopoImags \\ w/ intensity \& gradient filtrations (ours)\end{tabular}
        &\underline{87.07}	  & \textbf{87.20} 	  \\\hline\hline
    \end{tabular}
}
\end{table}
\begin{table}[t]
    \centering
    \caption{Results of various methods on the CBIS-DDSM dataset. 
    }
    \label{tab:CBIS-DDSM}
     \scalebox{0.8}{%
    \begin{tabular}{l|l|l}
        \hline
        Method &Acc ($\uparrow$)  &AUC ($\uparrow$) \\\hline
         PHG-Net~\cite{peng2024phg}&77.23	&83.39	\\\hline
        \hline
        ResNet50~\cite{he2016deep}&	70.60  & 75.80 	  \\
       \hline
       \begin{tabular}[c]{@{}l@{}} ResNet50 + PIs~\cite{adams2017persistence}\end{tabular}
      &	70.03  &  74.76	 \\
       \hline
       \begin{tabular}[c]{@{}l@{}} ResNet50 + LTEIs~\cite{kaji2020cubical}\end{tabular}
        &67.47	  & 73.96 	  \\\hline
        \begin{tabular}[c]{@{}l@{}} ResNet50 + TopoImage w/ intensity filtration\end{tabular}
        &71.31	  & 75.40	  \\\hline
         \begin{tabular}[c]{@{}l@{}} ResNet50 + TopoImage w/ gradient filtration\end{tabular}
        &71.74	  & 76.41  	  \\\hline
        \begin{tabular}[c]{@{}l@{}} ResNet50 + multi-view TopoImags \\ w/ intensity \& gradient filtrations (ours)\end{tabular}
        &73.03	  &  78.89	  \\\hline\hline
       ConvNeXt~\cite{liu2022convnet}&	75.28  & 80.28 	  \\
       \hline
       \begin{tabular}[c]{@{}l@{}} ConvNeXt + PIs\end{tabular}
      &	72.02  &  	78.92  \\
       \hline
       \begin{tabular}[c]{@{}l@{}} ConvNeXt + LTEIs\end{tabular}
        &	70.17  &  76.60	  \\\hline
        \begin{tabular}[c]{@{}l@{}} ConvNeXt + TopoImage w/ intensity filtration\end{tabular}
        &	76.16  &  80.77	  \\\hline
         \begin{tabular}[c]{@{}l@{}} ConvNeXt + TopoImage w/ gradient filtration\end{tabular}
        &	76.88  &  81.25	  \\\hline
        \begin{tabular}[c]{@{}l@{}} ConvNeXt + multi-view TopoImags \\ w/ intensity \& gradient filtrations (ours)\end{tabular}
        &	77.94  &  83.31	  \\\hline\hline
         Swin-Transformer v2~\cite{liu2022swin}&	73.60  &  79.86	  \\
       \hline
       \begin{tabular}[c]{@{}l@{}} Swin-Transformer v2 + PIs\end{tabular}
      &69.32	  &  	73.88  \\
       \hline
       \begin{tabular}[c]{@{}l@{}} Swin-Transformer v2 + LTEIs\end{tabular}
        &67.90	  &  71.70	  \\\hline
        \begin{tabular}[c]{@{}l@{}} Swin-Transformer v2 + TopoImage w/ intensity filtration\end{tabular}
        &	74.38  &  79.17	  \\\hline
         \begin{tabular}[c]{@{}l@{}} Swin-Transformer v2 + TopoImage w/ gradient filtration\end{tabular}
        &	74.74  & 80.48 	 \\\hline
        \begin{tabular}[c]{@{}l@{}} Swin-Transformer v2 + multi-view TopoImags \\ w/ intensity \& gradient filtrations (ours)\end{tabular}
        &	75.95  & 81.34 	  \\\hline\hline
           PHG-Net~\cite{peng2024phg}&78.22	  & 84.26 	  \\
       \hline
       \begin{tabular}[c]{@{}l@{}} PHG-Net + PIs\end{tabular}
      &	73.34  &  78.56 	  \\
       \hline
       \begin{tabular}[c]{@{}l@{}} PHG-Net + LTEIs\end{tabular}
        &	71.19  &  	 77.75  \\\hline
        \begin{tabular}[c]{@{}l@{}} PHG-Net + TopoImage w/ intensity filtration\end{tabular}
        &78.85	  &  84.11	  \\\hline
         \begin{tabular}[c]{@{}l@{}} PHG-Net + TopoImage w/ gradient filtration\end{tabular}
        &\underline{79.43}	  & \underline{84.45}	  \\\hline
        \begin{tabular}[c]{@{}l@{}} PHG-Net + multi-view TopoImags \\ w/ intensity \& gradient filtrations (ours)\end{tabular}
        &\textbf{80.01}  &\textbf{85.91} 	  \\\hline\hline
    \end{tabular}
}
\end{table}
\begin{table}[t]
    \centering
    \caption{Results of various methods on the ExtCRC dataset. 
    }
    \label{tab:extcrc}
     \scalebox{0.78}{%
    \begin{tabular}{l|l|l}
        \hline
        Method &Acc ($\uparrow$)  &AUC ($\uparrow$) \\\hline
         CA-CNN~\cite{shaban2020context}&86.70	&--	\\\hline
         CGC-Net~\cite{zhou2019cgc}&93.00	&--	\\
       \hline
        HAT-Net~\cite{su2021hat}&95.30	&--	\\
       \hline
        CCF-GNN~\cite{wang2023ccf}&96.60	&--	\\\hline
         Gu et al.~\cite{gu2024boosting} &83.52 	&92.69	\\
       \hline
        \hline
        ResNet50~\cite{he2016deep}&	95.00  &  99.20	  \\
       \hline
       \begin{tabular}[c]{@{}l@{}} ResNet50 + PIs~\cite{adams2017persistence}\end{tabular}
      &	96.33  &  	98.81  \\
       \hline
       \begin{tabular}[c]{@{}l@{}} ResNet50 + LTEIs~\cite{kaji2020cubical}\end{tabular}
        &	95.66  &  99.50	 \\\hline
        \begin{tabular}[c]{@{}l@{}} ResNet50 + TopoImage w/ intensity filtration\end{tabular}
        &	98.00  &  	99.55 \\\hline
         \begin{tabular}[c]{@{}l@{}} ResNet50 + TopoImage w/ gradient filtration\end{tabular}
        &	98.33  &  99.78	 \\\hline
        \begin{tabular}[c]{@{}l@{}} ResNet50 + multi-view TopoImags \\ w/ intensity \& gradient filtrations (ours)\end{tabular}
        &	98.66  &  99.68	  \\\hline\hline
       ConvNeXt~\cite{liu2022convnet}&96.34	  &  99.23	  \\
       \hline
       \begin{tabular}[c]{@{}l@{}} ConvNeXt + PIs\end{tabular}
      &89.67	  &  93.82	  \\
       \hline
       \begin{tabular}[c]{@{}l@{}} ConvNeXt + LTEIs\end{tabular}
        &	 87.00 &  93.75	  \\\hline
        \begin{tabular}[c]{@{}l@{}} ConvNeXt + TopoImage w/ intensity filtration\end{tabular}
        &	97.33  &  	99.61  \\\hline
         \begin{tabular}[c]{@{}l@{}} ConvNeXt + TopoImage w/ gradient filtration\end{tabular}
        &	\underline{98.67}  &  	99.57  \\\hline
        \begin{tabular}[c]{@{}l@{}} ConvNeXt + multi-view TopoImags \\ w/ intensity \& gradient filtrations (ours)\end{tabular}
        &	\textbf{99.00}  &  \underline{99.86}	  \\\hline\hline
         Swin-Transformer v2~\cite{liu2022swin}&	97.67  &  98.77	  \\
       \hline
       \begin{tabular}[c]{@{}l@{}} Swin-Transformer v2 + PIs\end{tabular}
      &93.33	  &  	97.95  \\
       \hline
       \begin{tabular}[c]{@{}l@{}} Swin-Transformer v2 + LTEIs\end{tabular}
        &91.67	  &  	97.14  \\\hline
        \begin{tabular}[c]{@{}l@{}} Swin-Transformer v2 + TopoImage w/ intensity filtration\end{tabular}
        &98.33	  & 99.04	  \\\hline
         \begin{tabular}[c]{@{}l@{}} Swin-Transformer v2 + TopoImage w/ gradient filtration\end{tabular}
        &	\underline{98.67}   &  99.39	  \\\hline
        \begin{tabular}[c]{@{}l@{}} Swin-Transformer v2 + multi-view TopoImags \\ w/ intensity \& gradient filtrations (ours)\end{tabular}
        &	\textbf{99.00}  &  \textbf{99.96}	  \\\hline\hline
      
    \end{tabular}
}
\end{table}

\subsection{Implementation Details}\label{imp}
Our experiments are conducted with the PyTorch framework. For computing 
PDs of patches, we employ the Cubical Ripser software.
To transform PDs into PIs, we utilize the Persim package.
Model training is performed on an NVIDIA Tesla V100 Graphics Card with 32GB GPU memory using the AdamW optimizer~\cite{loshchilov2017decoupled} with a weight decay = $0.005$. 
The learning rate is 0.0001, and the number of training epochs is 400 for the experiments. The batch size for each case is set as the maximum size allowed by the GPU.
We apply standard data augmentation, such as rotation, random flip, random crop, etc., to avoid overfitting.

\subsection{Experimental Results}\label{exp-results}
\textbf{ISIC 2017 Results.}
To evaluate our method, we use two prominent CNN-based (ResNet50~\cite{he2016deep} and ConvNeXt \cite{liu2022convnet}), one of the latest Transformer-based (Swin-Transformer v2~\cite{liu2022swin}), and a recent topology-based (PHG-Net~\cite{peng2024phg}) DL classification networks.
%
We first compute the corresponding TopoImages of the input images using the intensity and gradient filtration functions for the dataset with Algorithm~\ref{algorithm1}.
We then complement the input images with the TopoImages (generated using a single filtration function) and multi-view TopoImages using our fusion module (see Fig.~\ref{fig:pipeline}(b)).
Finally, we train these four models with the fused images.
We compare our method
against SOTA DL models (i.e., D{\'\i}az~\cite{diaz2017incorporating},  ARDT-DenseNet~\cite{wu2020skin}, and Gu et al.~\cite{gu2024boosting}) and two vectorization methods (i.e., PIs~\cite{adams2017persistence} and lifetime-enhanced images (LTEIs)~\cite{kaji2020cubical}).

Table~\ref{tab:isic17} presents the results. We can observe that 
(i) The models trained with PIs and LTEIs generally show a drop in AUC scores, except for a slight improvement when ResNet50 is trained with PIs.
In contrast, the models trained with our fused images (i.e., complementing input images with TopoImages) demonstrate considerable performance improvements. This is attributed to the encoded local topology information within the TopoImages.
(ii) When the input images are complemented with multi-view TopoImages, the performance is further enhanced by 2.46\%, 2.43\%, 1.93\%, and 2.93\% on ResNet50, ConvNeXt, Swin-Transformer v2, and PHG-Net, respectively. These results validate the effectiveness of our method in improving the performances of medical image classification tasks.
(iii) When PHG-Net (a global topology-based method) is combined with multi-view TopoImages, an additional gain is observed, demonstrating that our local topology-based approach is complementary to PHG-Net’s global topology-based method.

\textbf{CBIS-DDSM Results.}
We evaluate our method using the same four DL models as on the ISIC 2017 dataset: ResNet50, ConvNeXt, Swin-Transformer v2, and PHG-Net. 
Similarly, we complement input images with TopoImages and multi-view TopoImages, and train these models with the fused images.
%
Table~\ref{tab:CBIS-DDSM} reports the results, revealing similar observations as those made on the ISIC 2017 dataset: complementing input images with TopoImages results in considerable performance improvements, and the performance is further enhanced when input images are complemented with multi-view TopoImages, yielding accuracy improvements of 2.43\%, 2.66\%, 2.35\%, and 1.79\% on the four models
respectively.
The results demonstrate the effectiveness of our method in combining local topology encoding with multiple filtration functions to improve performances of medical image classification tasks.

\textbf{ExtCRC Results.}
Table~\ref{tab:extcrc} reports the results on the ExtCRC dataset. We can observe the following.
(i) The baselines, ResNet50, ConvNeXt, and Swin-Transformer v2, already achieve very high Accuracy performances, 95.00\%, 96.34\%, and 97.67\%, respectively, giving a very limited margin for big improvement.
Still, the models trained with our fused images (i.e., complementing input images with multi-view TopoImages) still manage to obtain
considerable performance improvements, confirming the
effectiveness of our new method. Specifically, ResNet50’s Accuracy is improved by 3.66\%,  ConvNeXt's is improved by 2.66\%, and Swin-Transformer v2’s is improved by 1.33\%.
(ii) One can see that the previous best-known method on this dataset is CCF-GNN~\cite{wang2023ccf}, whose Accuracy is
higher than the previous second-best method, HAT-Net~\cite{su2021hat}, by
a margin of 1.3\%. In comparison, our method outperforms
CCF-GNN by 2.40\% and HAT-Net by a notable 3.70\% in Accuracy, achieving new SOTA performance.
These impressive improvements validate the effectiveness of our method.

\subsection{Effect of Patch Size Choices}\label{sub:patch-size}
Our method is designed to encode local topological information in patches of input images.
It is important to investigate how the choice of patch size may affect classification performance.
Hence, we conduct experiments to examine the impact of different patch sizes.
Since an input image has 224$\times$224 pixels, we consider patch sizes that are reduced repeatedly by half of the input image's size, including 7$\times$7, 14$\times$14, 28$\times$28, 56$\times$56, and 112$\times$112. 
Specifically, we train the ResNet50 model using fused images (i.e., complementing input images with multi-view TopoImages) from the ISIC 2017, CBIS-DDSM, and ExtCRC datasets.
As shown in Table~\ref{tab:patch-size}, the ResNet50 model trained with multi-view TopoImages generated with 112$\times$112, 56$\times$56, and 28$\times$28 patches exhibited the best performances on the ISIC 2017, CBIS-DDSM, and ExtCRC datasets, respectively.

In our method, we use SAM~\cite{kirillov2023segment} to help find the approximate average size of objects in each dataset. Specifically, the approximate average sizes of objects in the ISIC 2017, CBIS-DDSM, and ExtCRC datasets are 11,025, 4,019, and 587 pixels, respectively, which closely align with 112$\times$112, 56$\times$56, and 28$\times$28 patch sizes, respectively. 
This suggests that the effective patch size should align with the average size of objects in the images.
We note that our primary motivation for using SAM is to estimate object sizes without requiring manual annotation. This SAM-based size estimation is performed only once, offline and label-free, and it is not required during inference or training, adding only a minimal extra cost (approximately 1.1 hours on ISIC 2017 dataset using a single NVIDIA Tesla V100 GPU).

\begin{table}[t!]
\centering
\caption{Results of using different patch sizes on the three datasets.}\label{tab:patch-size}
\centering
 \scalebox{0.88}{%
\begin{tabular}{l|c|c|c}
\multirow{3}{*}{Patch Size}  & \multicolumn{3}{c}{Datasets}  \\
 &{\begin{tabular}[c]{@{}l@{}} ISIC 2017 (AUC)\end{tabular}} & \begin{tabular}[c]{@{}l@{}} CBIS-DDSM (Acc)\end{tabular} & \begin{tabular}[c]{@{}l@{}} ExtCRC (Acc)\end{tabular} \\
\hline
7 $\times$  7  & 74.92 &67.61 & 91.67  \\\hline
14 $\times$  14  &76.74  &70.31 & 96.67  \\\hline
28 $\times$  28  &79.61  &\underline{71.31} &  \textbf{98.66} \\\hline
56 $\times$  56  & 80.29 &\textbf{73.03} &  \underline{98.33}\\\hline
112 $\times$  112  &\textbf{81.47}  &70.74 &  98.00 \\\hline
224 $\times$  224  & \underline{80.88} &70.60 &  95.00 \\\hline
\end{tabular}	
}
\end{table}


\subsection{Effect of Fusion Choices}\label{sub:fusion}
We conduct experiments to investigate the effect of different fusion methods. We consider the following fusion methods: (1) Directly concatenating multi-view TopoImages with input images, (2) directly mean pooling multi-view TopoImages with input images, and (3) fusing multi-view TopoImages with input images using the proposed CMVFM (see Fig.~\ref{fig:pipeline}(b)). Specifically, we train the ConvNeXt and Swin-Transformer v2 models using fused images from the CBIS-DDSM dataset. As shown in Table~\ref{tab:fusion-meth}, directly concatenating and applying mean pooling to multi-view TopoImages with the input images leads to a performance decline, whereas complementing input images with multi-view TopoImages using the proposed CMVFM results in performance improvements.
This validates the effectiveness of our proposed CMVFM.
%

\begin{table}[t]
    \centering
    \caption{Results of fusion methods on the CBIS-DDSM dataset. 
    }
    \label{tab:fusion-meth}
    \scalebox{0.8}{%
    \begin{tabular}{l|l}
        \hline
        Method &Acc ($\uparrow$)  \\\hline\hline
        ConvNeXt~\cite{liu2022convnet}&	75.28	\\
       \hline
       \begin{tabular}[c]{@{}l@{}} ConvNeXt + multi-view TopoImages w/ concatenation\end{tabular}
          &	73.26	\\\hline
          \begin{tabular}[c]{@{}l@{}} ConvNeXt + multi-view TopoImages w/ mean pooling\end{tabular}
          &	68.44	\\\hline
         \begin{tabular}[c]{@{}l@{}} ConvNeXt + multi-view TopoImages w/ CMVFM\end{tabular}
          &	\textbf{77.94} 	\\
       \hline
       \hline
         Swin-Transformer v2~\cite{liu2022swin}&		73.60\\
       \hline
       \begin{tabular}[c]{@{}l@{}} Swin-Transformer v2 + multi-view TopoImages w/ concatenation\end{tabular}
          &	71.49	\\\hline
           \begin{tabular}[c]{@{}l@{}} Swin-Transformer v2 + multi-view TopoImages w/ mean pooling\end{tabular}
          &	69.03	\\\hline
         \begin{tabular}[c]{@{}l@{}} Swin-Transformer v2 + multi-view TopoImages w/ CMVFM\end{tabular}
          &	\underline{75.95} 	\\
       \hline
       \hline
    \end{tabular}
}
\end{table}


\subsection{Computational Complexity of TopoImages
}\label{sub:time}
Our TopoImages computation is conducted on an NVIDIA Tesla V100 Graphics Card with 32GB GPU memory. Here, we report the time complexity of one dataset. 
It takes 11 minutes and 29 seconds to compute the corresponding TopoImages (using the intensity filtration function) on the ISIC 2017 dataset (2,600 images, each 224×224 pixels) when dividing the images into 112×112 size patches.
The TopoImages computation time on smaller datasets may be significantly shorter, and the use of multiple GPUs could further accelerate the TopoImages computation.

\begin{figure}[t!]
    \centering
    \includegraphics[width=0.88\columnwidth]{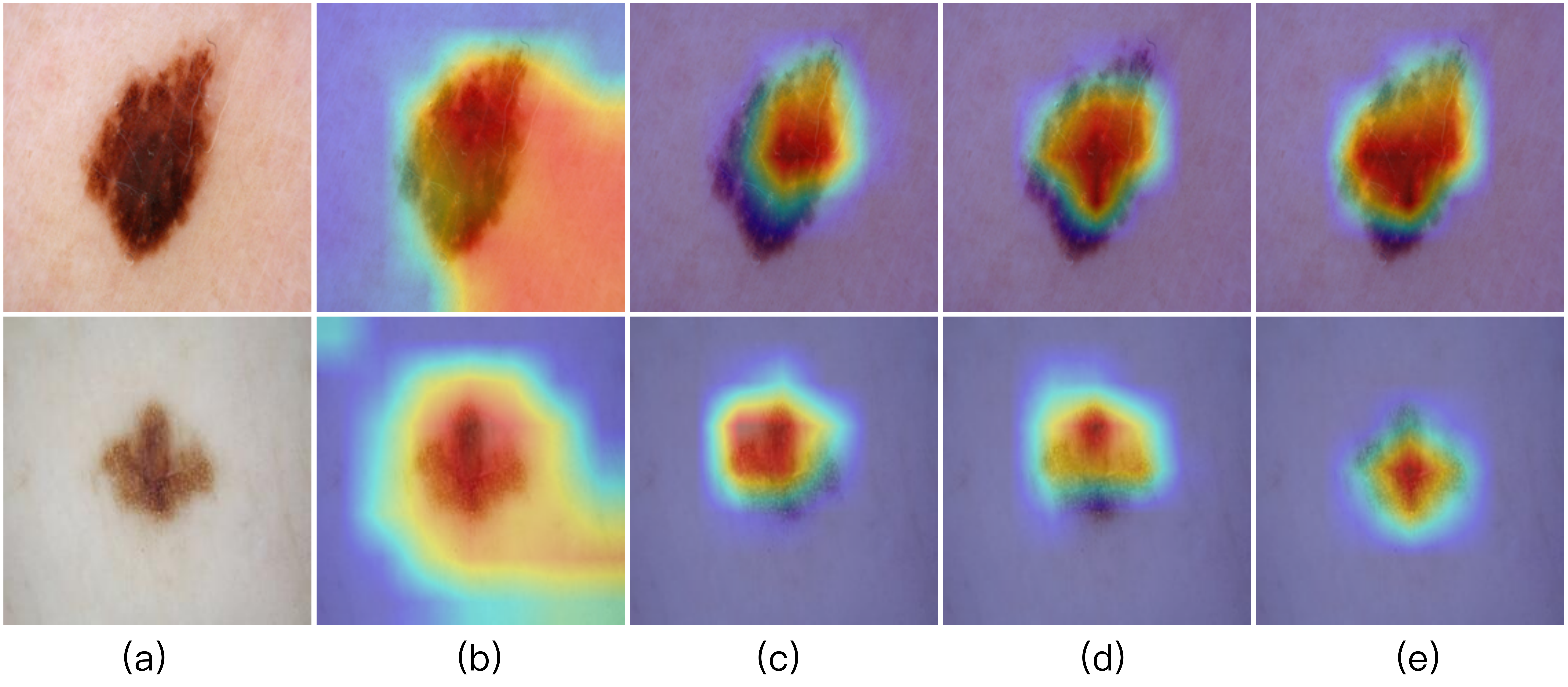}
    \caption{Qualitative examples of different filtration methods on
    the ISIC 2017 dataset~\cite{codella2018skin}. (a) Input image; (b) original image without TopoImage; (c) TopoImage with intensity filtration; (d) TopoImage with gradient filtration; and (e) multi-view TopoImags with intensity \& gradient filtrations (ours).
    } 
    \label{fig-vis}
\end{figure}

\subsection{Qualitative Results
}\label{sub:qualitative}
Fig.~\ref{fig-vis} shows the heatmaps of different methods. Specifically, we train the ConvNeXt\cite{liu2022convnet} model using original images, fused images that merge original images with TopoImages using intensity filtration, fused images that merge original images with TopoImages using gradient filtration, and fused images that merge original images with multi-view TopoImages using both intensity and gradient filtrations.
From the visual results, we observe the following: (1) Fusing original images with TopoImages using intensity or gradient filtration helps the model concentrate on the relevant regions; (2) Fusing original images with multi-view TopoImages using both intensity and gradient filtrations significantly enhances the model’s focus on the correct regions in the images.

Local topological features, such as loops ($1$D holes), connectivity patterns ($0$D connected components), and small cavities, can highlight subtle morphological changes in specific regions of an image. In contrast, global topological descriptors may overlook these finer details when large, dominant structures are present. As illustrated in Fig.~\ref{fig-vis}, local loops around gland or lesion boundaries can expose boundary irregularities, while small holes in tissue regions may indicate pathological changes. Our patch-based local topological encoding is designed to capture these local abnormalities (e.g., irregular gland shapes or melanoma lesion edges), which are highly relevant to accurate classification.

\section{Conclusions}
\label{sec:Conclusions}
In this paper, we proposed a new general approach, TopoImages, for computing a new representation of medical images by encoding local topology information of patches in input images using multiple filtration functions.
Our multi-view TopoImages have the same formation (i.e., height and width) as the input images, enabling them to be seamlessly integrated into existing DL models without modifying their architectures.
Experiments on three public medical image datasets verified the effectiveness of our TopoImages on classification tasks.
We expect that our TopoImages approach will offer a new perspective for data representation and open new avenues for capturing topological information in complex medical image datasets for various applications. 

\section{Acknowledgments}
\label{sec:acknowledgements}
This research was supported in part by NSF Grant OAC-2104158. 

\bibliographystyle{ACM-Reference-Format}
\bibliography{sample-base}

\end{document}